\pdfoutput=1
\documentclass[10pt,journal,compsoc]{IEEEtran}
\usepackage{cite}
\usepackage{multirow}
\usepackage{graphicx}
\usepackage{color}
\usepackage{dcolumn}
\usepackage{amsmath}
\usepackage{kantlipsum}
\usepackage{url}
\usepackage{microtype}
\usepackage{verbatim}
\usepackage{colortbl}
\usepackage{arydshln}
\usepackage{multirow}
\newcolumntype{K}[1]{>{\centering\arraybackslash}p{#1}}
\usepackage{booktabs}
\usepackage[flushleft]{threeparttable}
\usepackage{hhline}
\usepackage{algorithm}
\usepackage{algorithmic}
\usepackage{mathtools}
\usepackage{arydshln}

\ifCLASSOPTIONcompsoc
\usepackage[caption=false, font=normalsize, labelfont=sf, textfont=sf]{subfig}
\else
\usepackage[caption=false, font=footnotesize]{subfig}
\fi
\begin{document}
\title{STEERAGE: \textbf{S}yn\textbf{t}hesis of N\textbf{e}ural N\textbf{e}tworks Using A\textbf{r}chitecture Se\textbf{a}rch and \textbf{G}row-and-Prun\textbf{e} Methods}

\author{Shayan~Hassantabar,~Xiaoliang~Dai,~and~Niraj~K.~Jha,~\IEEEmembership{Fellow,~IEEE}
\thanks{This work was supported in part by NSF Grant No. CNS-1617640 and in part by
NSF Grant No. CNS-1907381.
Shayan Hassantabar, Xiaoliang Dai, and Niraj K. Jha are with the Department
of Electrical Engineering, Princeton University, Princeton,
NJ, 08544 USA, e-mail:\{seyedh, xdai,jha\}@princeton.edu.}}

\IEEEtitleabstractindextext{%
\begin{abstract}
Neural networks (NNs) have been successfully deployed in various applications 
of Artificial Intelligence.  However, architectural design of these models is 
still a challenging problem. This is due to the need to navigate a large 
number of hyperparameters that forces the search space of possible 
architectures to grow exponentially. Furthermore, using a trial-and-error 
design approach is very time-consuming and leads to suboptimal architectures.  
In addition, neural networks are known to have a lot of redundancy. This 
increases the computational cost of inference and poses a severe obstacle to
deployment on Internet-of-Thing (IoT) sensors and edge devices.  To address 
these challenges, we propose the STEERAGE synthesis methodology.  It consists 
of two complementary approaches: intelligent and efficient architecture 
search, and grow-and-prune NN synthesis.  The first step, incorporated in a 
global search module, uses an accuracy predictor to efficiently navigate the 
architectural search space.  This predictor is built using boosted 
decision tree regression, iterative sampling, and efficient evolutionary 
search.  This step starts from a base architecture and explores the 
architecture search space to obtain a variant of the base architecture with 
the highest performance.  The second step involves local search. By taking 
advantage of various grow-and-prune methodologies for synthesizing 
convolutional and feed-forward NNs, it not only reduces network redundancy and 
computational cost, but also boosts model performance.  We have evaluated 
STEERAGE performance on various datasets, including MNIST and CIFAR-$10$. We 
demonstrate significant accuracy improvements over the baseline architectures.  For the MNIST dataset, our CNN architecture achieves an error rate 
of $0.66\%$, with $8.6\times$ fewer parameters compared to the LeNet-$5$ 
baseline.  For the CIFAR-$10$ dataset, we used the ResNet architectures as the 
baseline. Our STEERAGE-synthesized ResNet-$18$ has a $2.52\%$ accuracy 
improvement over the original ResNet-$18$, $1.74\%$ over ResNet-$101$, and 0.16\% over
ResNet-1001, while requiring the number of parameters and floating-point operations
comparable to the original ResNet-18. This demonstrates that instead of just increasing the
number of layers to increase accuracy, an alternative is to use a better NN architecture
with a small number of layers.
In addition, STEERAGE achieves an error rate of just $3.86\%$ with a variant of 
ResNet architecture with $40$ layers. To the best of our knowledge, this is 
the highest accuracy obtained by ResNet-based architectures on the 
CIFAR-$10$ dataset.  STEERAGE also obtains the highest accuracy for
various other feedforward NNs geared towards edge devices and IoT sensors.
\end{abstract}

}
\maketitle

\IEEEdisplaynontitleabstractindextext

\IEEEpeerreviewmaketitle

\IEEEraisesectionheading{\section{Introduction}\label{sec:introduction}}

\IEEEPARstart{N}{eural} networks (NNs) have led to wonderful achievements in various areas of 
Artificial Intelligence, such as computer vision \cite{krizhevsky2012imagenet}, speech recognition 
\cite{graves2013speech, kim2017residual}, and machine translation \cite{jean2014using}. Deep 
NNs can learn hierarchical features, which is key to the above-mentioned triumphs. Improving 
the performance of these models is a very active current line of research \cite{akmandor2019secret}.

The NN architecture affects the performance of the final model. As a result, synthesizing an
appropriate NN architecture is of utmost importance. However, since the NN architecture search space 
is exponentially large in its many hyperparameters, using a trial-and-error approach to
constructing the NN leads to suboptimal results. Hence, it is necessary to find the best set of 
hyperparameter values for the architecture. 

Another important design objective is model compactness. Compact models are easier to deploy on edge 
devices. As shown previously \cite{han2015learning, hassantabar2019scann}, the number of NN 
parameters can be reduced significantly without degrading performance.  Thus, choosing the
hyperparameter values that yield very compact, not just accurate, NN architectures is also an
important consideration. 

Neural architecture search (NAS) is a technique for automatically synthesizing convolutional layers 
through hyperparameter search.  It is currently a very active area of research 
\cite{baker2016designing, tan2018mnasnet}. However, most NAS approaches, such as those based on 
reinforcement learning (RL) \cite{baker2016designing, liu2018progressive}, are computationally
very expensive.

To address the above challenges, we propose a two-step global+local search based NN synthesis 
methodology, called STEERAGE \footnote{Steerage refers to the action of steering a boat. Here, we use 
it to steer NN synthesis.} (\textbf{S}yn\textbf{T}hesis of n\textbf{E}ural n\textbf{E}tworks with 
a\textbf{R}chitecture se\textbf{A}rch and \textbf{G}row and prun\textbf{E} methods). STEERAGE is 
efficient and automatically generates accurate and compact NN architectures. To efficiently find the 
NN hyperparameter values, we derive an accuracy predictor to measure architecture fitness. 
This significantly speeds up architecture search.
We use grow-and-prune synthesis methods to ensure model compactness. 
STEERAGE targets both the convolutional and feed-forward layers of the NN through two sets of
hyperparameter values, and alleviates the curse of dimensionality by incorporating dimensionality 
reduction (DR) in the search space.

Global search is based on the NAS approach.  However, unlike prior work, it does not rely on 
especially-crafted efficient building blocks, nor does it use the computationally expensive 
RL approach. It uses an accuracy predictor to measure architecture performance.  This obviates
the need to train the NNs during search, thus significantly speeding up the search process. We use 
an iterative process to build the accuracy predictor, using a combination of a boosted 
decision tree regression
tree and quasi Monte-Carlo (QMC) sampling. We employ an efficient evolutionary search (EES)
method to find the best-performing variant of the base architecture in the search space.  

The second synthesis step, local search, starts with the best architecture obtained from global 
search.  It not only enhances model performance, but also generates a compact NN by reducing 
architectural redundancy.  We use two different grow-and-prune methodologies for this purpose:
targeted at feed-forward NNs (FFNNs) and convolutional NNs (CNNs). For FFNNs, we use 
the SCANN \cite{hassantabar2019scann} synthesis methodology to enhance compactness and accuracy. 
SCANN uses three operations iteratively: connection growth, neuron growth, and connection pruning. 
We use the NeST \cite{dai2017nest} synthesis methodology for the convolutional layers. 
It uses two operations: feature map growth and partial-area convolution. 

Our contributions can be summarized as follows:
\begin{itemize}
    \item We present STEERAGE, an NN synthesis methodology that efficiently explores both
the FFNN and CNN search spaces for edge applications.
    \item STEERAGE employs a two-step synthesis approach that utilizes two complementary approaches: 
architecture search and network growth-and-pruning. The first step uses an accuracy predictor based 
on boosted decision tree regression for efficient architecture search. The second step helps 
refine the architecture to improve its performance and reduce redundancy.
    \item STEERAGE is general and easy-to-extend, and adapts to various architecture types, such as 
FFNNs as well as shallow and deep CNNs.
    \item We demonstrate the effectiveness of STEERAGE through evaluation on
$11$ datasets, including MNIST and CIFAR-$10$.
\end{itemize}{}

The rest of the article is organized as follows. Section \ref{sect:related} presents
related work. Section \ref{sect:background} provides the necessary background material.
Section \ref{sect:synthesis} discusses the STEERAGE synthesis methodology, and the complementary 
nature of its global and local search modules. Section \ref{sect:expresults} presents experimental 
results. Section \ref{sect:discussion} provides a short discussion of the methodology and results. 
Finally, Section \ref{sect:conclusion} concludes the article.

\section{Related Work}
\label{sect:related}
In this section, we summarize prior work on NN synthesis.

\subsection{Synthesizing Compact Architectures}
Synthesis of compact NNs through removal of redundant connections and neurons has attracted
significant attention. Network pruning is one such widely used approach 
\cite{han2015learning,dai2017nest, molchanov2016pruning,yang2018netadapt, dai2018grow, 
hassantabar2019scann, dai2019incremental}. Several works use structured sparsity to prune deep 
CNNs \cite{wen2016learning, ye2018rethinking}.  RL has also been used to compress CNN models for 
mobile devices \cite{he2018amc}. Runtime pruning \cite{lin2017runtime} separately uses RL to evaluate 
the importance of each convolutional feature map and channel pruning for each input image. 
Dynamic channel pruning \cite{gao2018dynamic} is another approach for pruning channels based on 
their relevance at runtime. 

Another approach for generating compact CNNs is to handcraft efficient building blocks. It is
aimed at NN deployment on mobile devices. One operation that significantly reduces computational cost 
is depthwise separable convolution. It has been used in both MobileNet architecture 
versions \cite{howard2017mobilenets, sandler2018mobilenetv2}. MobileNetV$2$ 
\cite{sandler2018mobilenetv2} uses efficient CNN blocks with inverted residuals and linear 
bottlenecks. Other such operations are pointwise group convolution and channel shuffle that are used 
in the ShuffleNet architectures \cite{zhang2018shufflenet, ma2018shufflenet}. 
In addition, quantization can also reduce the computational cost of inference \cite{han2015deep, 
hubara2016binarized, zhu2016trained}, with a minimal adverse impact on  performance.

\subsection{Neural Architecture Search (NAS)}

RL is a well-known method for synthesizing NNs \cite{baker2016designing, liu2018progressive, 
pham2018efficient}.  MnasNet \cite{tan2018mnasnet} uses a platform-aware NAS approach based on RL, 
built atop MobileNetV$2$. It uses different layer structures for different parts of the network. 
MobileNetV$3$ uses two complementary search approaches. It combines NAS for block-wise search and 
NetAdapt \cite{yang2018netadapt} for layer-wise search (for individual fine-tuning of the layers). 
Another NAS approach is weight sharing \cite{brock2017smash, bender2018understanding }. It
trains an over-parametrized supernet architecture. The differentiable NAS approach uses a loss 
function to optimize the weights and architectural parameters \cite{liu2018darts, veniat2018learning, 
xie2018snas, wu2019fbnet, guo2019single}. The final architecture inherits the weights of the supernet 
and is then fine-tuned. 

\section{Background}
\label{sect:background}
In this section, we familiarize the reader with some necessary background material. First, we give an 
overview of the SCANN \cite{hassantabar2019scann} synthesis framework since it is used within 
STEERAGE.  We also discuss the Chamnet \cite{dai2019chamnet} framework that first introduced
the idea of an accuracy predictor to expedite NN synthesis.

\subsection{SCANN}
In this section, we first briefly explain the operations used in SCANN, which are used to change 
the NN architecture in different steps of the training process. Then, we go over three different training 
schemes employed in this framework. 

SCANN uses three main operations to change the network architecture, namely, connection growth, 
neuron growth, and connection pruning. Connection growth adds a number of inactive connections to 
the network. Connections may be added based on their effect on the loss function. 
Neuron growth works either by duplicating existing neurons in the network or by randomly generating 
new neurons and their connections. Connection pruning reduces the size of the network. This is done 
by either pruning small-weight or large-weight connections. 

The above operations are used in three different training schemes. After applying any of the 
architecture-changing operations, the weights of the network are trained and the performance of the 
architecture evaluated on the validation set. The three different training schemes are referred 
to as Schemes A, B, and C. Scheme A starts with a small network and gradually increases its size. 
On the other hand, Scheme B starts with a large network and gradually decreases its size. Scheme C 
is similar to Scheme B, but limits the NN to a multi-layer perceptron (MLP) architecture by only 
allowing connections to adjacent layers. 

\subsection{ChamNet}
ChamNet is aimed at efficiently adapting a given NN to a resource-constrained computational platform.
It uses three different predictors for this purpose: accuracy, latency, and energy consumption of 
the NN.  With the help of these predictors, model adaptation is framed as a constraint optimization 
problem that aims to maximize accuracy while satisfying the energy and latency constraints of the 
platform. 

Chamnet uses Gaussian process (GP) regression to train the accuracy predictor.  The latency
predictor is based on an operator latency look-up table. The energy predictor 
is also built atop GP regression, with energy consumption values directly measured from the hardware 
platform.  These three predictors are used in an efficient NN search process (random and genetic 
search) to solve the constrained optimization problem. 

\section{The STEERAGE Synthesis Framework}
\label{sect:synthesis}
In this section, we first give an overview of the STEERAGE synthesis framework. We then zoom into 
the search space and introduce the DR methods that we used as one of the search 
dimensions. We then go on to describe the complete list of search dimensions (hyperparameters) for both CNNs and FFNNs. 
This is followed by the description of our accuracy predictor and search mechanisms. 

\subsection{Framework Overview}
\begin{figure*}[!ht]
    \centering
    \includegraphics[scale=0.22]{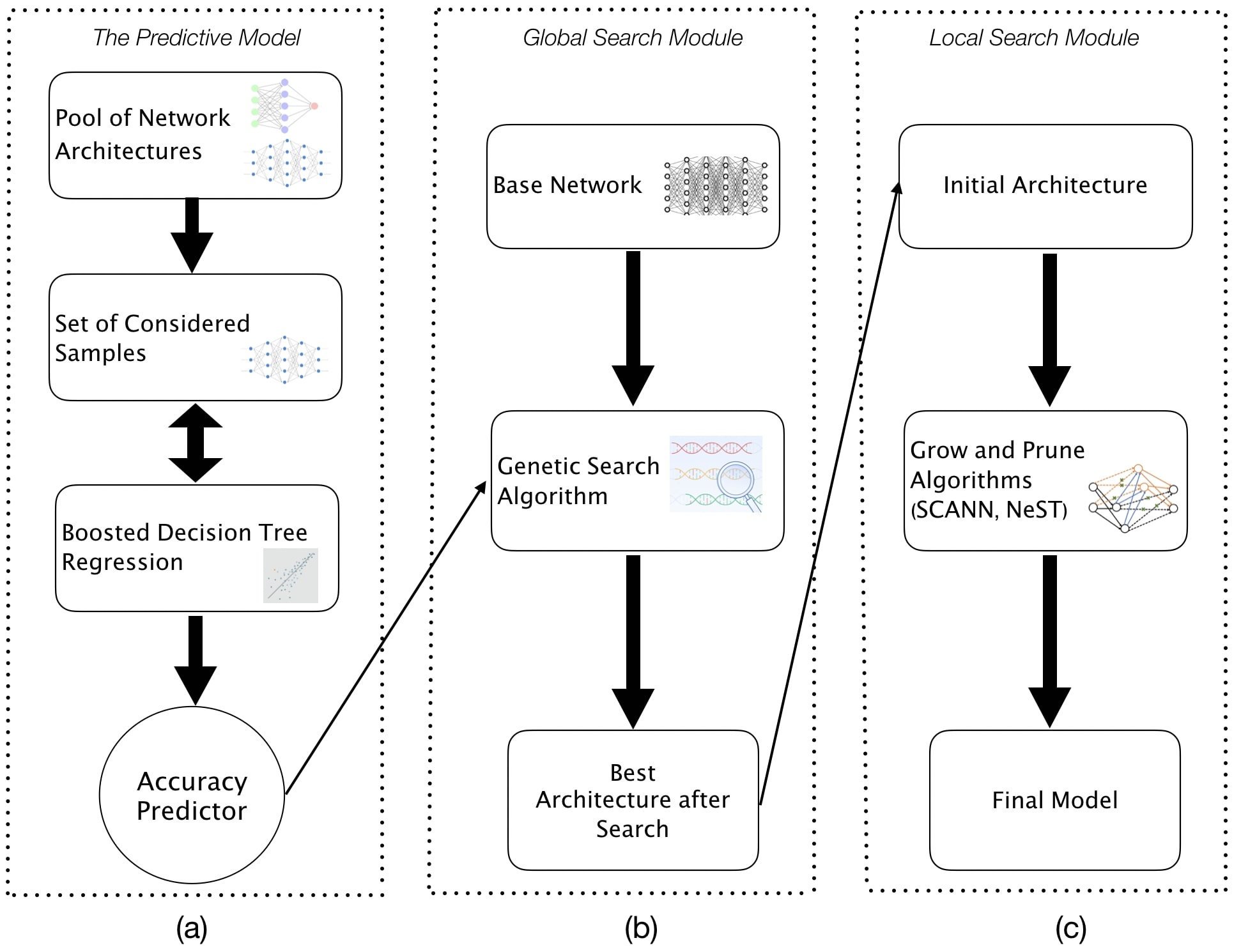}
    \caption{Block Diagram of STEERAGE: (a) Training an accuracy predictor, (b) genetic search to 
find the architecture with maximum accuracy, and (c) local search around the architecture 
through efficient grow-and-prune algorithms.}
\label{fig:diagram}
\end{figure*}

Fig.~\ref{fig:diagram} shows the block diagram of the STEERAGE methodology.  It takes the base 
network and dataset as input and, based on an evolutionary search and network grow-and-prune
algorithms, finds a variant of the base NN that is superior in terms of classification performance. 
For evolutionary search, we represent each architecture with a vector of hyperparameters 
$\textbf{x} \in \mathcal{R}^n$, where $n$ is the number of hyperparameters in the search space. 

We evaluate the fitness of each vector representation of the architecture by using an accuracy 
predictor.  This predictor is trained using a number of sample architectures that are chosen 
iteratively.  During training, the accuracy of the architectures is used as their measure of fitness. 
EES then selects the fittest candidates to generate the next round of architectures using mutation 
and crossover operations. We formulate evolutionary search as an optimization problem that maximizes 
the predicted accuracy of the architecture:
\[
\text{Acc}(x) = \text{Predicted-Acc} (x)
\]
\[
\textbf{x} = \textit{argmax} (\text{Acc})
\] 

The accuracy predictor is used to find the gene $x$ that maximizes accuracy.
The process of building this predictor is discussed in Section~\ref{sect:acc-pred}. 

As can be seen from Fig.~\ref{fig:diagram}, the candidate architecture output of EES is fed to the
local search module. In this step, we use two different grow-and-prune synthesis methodologies,
SCANN and NeST, to synthesize the feed-forward and convolutional parts of the network, respectively. 
This step finds the best NN architecture, given the output of the EES step. This process is described 
in Section \ref{sect:grow-and-prune}. 

\subsection{Dimensionality Reduction}
\label{sect:dr}
Curse of dimensionality is a well-known problem that stems from the need to increase the
dataset size as its dimensionality increases.  Since it is often not possible to increase
dataset size, researchers traditionally use various DR techniques. 
Since DR methods are quite effective, we include DR as one of the hyperparameters in our search 
space. It enables us to reduce the number of features and map the $N \times d$-dimensional dataset 
to a $N \times k$-dimensional dataset, where $k < d$.  We use two different sets of DR methods 
for FFNNs and CNNs, as described next.

\subsubsection{FFNNs}
The framework chooses amongst $11$ different DR methods for FFNNs. Random projection (RP) is a 
DR method based on the Johnson-Lindenstrauss lemma \cite{sivakumar2002algorithmic, 
dasgupta2003elementary}. We consider four different random projection matrices that yield four 
different DR methods. The entries in two of the matrices are i.i.d. samples drawn from the Gaussian 
distributions, $\mathcal{N}(0,\frac{1}{k})$ and $\mathcal{N}(0,1)$. Entries of the other two matrices 
are drawn from the following probability distributions \cite{achlioptas2001database}:

$$
\phi_{ij} = \begin{cases}
+1 & \text{with probability   }  \frac{1}{2}\\
-1 & \text{with probability   } \frac{1}{2}\\
\end{cases}
$$

$$
    \phi_{ij} = \sqrt{\frac{3}{k}} \begin{cases}
    1 & \text{with probability  } \frac{1}{6}\\
    0 & \text{with probability  } \frac{2}{3}\\
    -1 & \text{with probability  } \frac{1}{6}\\
    \end{cases}
$$

We also include a number of traditional DR methods in the set of choices to the framework: 
principal component analysis (PCA), polynomial kernel PCA, Gaussian kernel PCA, factor analysis, 
isomap, independent component analysis, and spectral embedding. We used the scikit-learn machine 
learning library \cite{pedregosa2011scikit} implementation of these methods.

\subsubsection{CNNs}
For image datasets that are analyzed by CNN architectures, we use different downsampling methods. 
Similar to the methods used in \cite{chrabaszcz2017downsampled}, we use 
lanczos, nearest, bilinear, bicubic, Hamming, and box filter from the \textit{Pillow} library \cite{pillow}.

Lanczos uses a low-pass filter to smoothly interpolate the pixel values in the resampled image. 
Nearest neighbor interpolation, used in the nearest filter, is a simple method that only selects the 
value of the nearest neighbor to do the interpolation, and produces a piece-wise constant interpolant. 
Bilinear interpolation is an extension of linear interpolation based on quadratic functions. 
The bicubic filter is an extension of cubic interpolation and is based on Lagrange polynomials, cubic 
splines, and the cubic convolution algorithm.  The Hamming filter is similar to bilinear, but 
generates sharper images.  
The box filter uses nearest neighbor interpolation for upscaling. Furthermore, each pixel of the original image contributes to one pixel of the resampled image with identical weights. 

\begin{figure}[!h]
    \centering
    \includegraphics[scale=0.25]{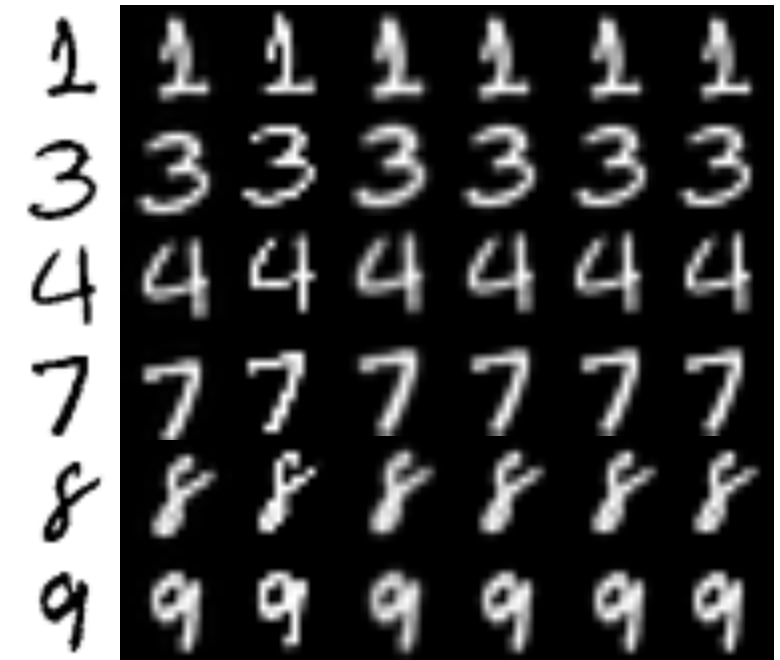}
    \caption{Original MNIST (size $28\times28$) (left column), and the downsampled (size $14\times14$) versions using six different techniques: lanczos, nearest, bilinear, bicubic, Hamming, and box filter, from left to right, 
respectively.}
    \label{fig:cnn-sample}
\end{figure}

Using the above methods, we downsample the original $28 \times 28$ size images of the MNIST dataset to 
$21 \times 21$ and $14 \times 14$ size images. Similarly, we reduce the $32 \times 32$ size images of 
the CIFAR-$10$ dataset to $24 \times 24$ size images.  Fig.~\ref{fig:cnn-sample} shows examples of 
the downsampled images from the MNIST dataset using these methods.

\subsection{Search Vector Space}
The set of hyperparameters chosen to navigate the search space directly impacts the performance of the
STEERAGE framework. Hence, using important design hyperparameters to define the search space is 
necessary.  However, including too many hyperparamters may have a negative impact. For example, 
as the number of search hyperparameters increases, we need more data to train the accuracy predictor. 
Furthermore, including architectures with inferior performance can also hurt STEERAGE performance. 
Therefore, the set of hyperparameters should be as small as possible, yet be inclusive of all 
important factors. 

We differentiate between the set of hyperparameters chosen for FFNNs and CNNs. These sets are
discussed next.

\subsubsection{FFNNs}
For FFNNs, we use five hyperparameters to define the search space. These are
(i) the number of hidden layers in the architecture, (ii) the number of neurons in each layer, 
(iii) the DR method, (iv) the feature compression ratio of the DR method, and (v) network 
quantization.  Quantization is done in the inference step: we considered various bit-length 
representations of the network weights (e.g., $4$, $8$, or $16$ bits).

\subsubsection{CNNs}
There are a number of hyperparameters that significantly impact the performance of CNNs. 
They include the number of convolutional layers, the number of feature maps in each layer, 
and the filter size of each layer. Pooling layers are often used to reduce the spatial size of the 
feature maps in the convolutional layers. The use of average pooling or max-pooling can, hence, be 
another hyperparameter included in the search space. We also include some hyperparameters 
that are similar to those targeted for FFNNs, e.g., the DR method, the number of hidden 
layers, and the number of neurons in each layer of the fully-connected classifier, and quantization 
bit representation.

\subsection{Building the Accuracy Predictor}
\label{sect:acc-pred}
In order to facilitate and accelerate architecture search, we need an efficient way to evaluate the 
fitness of each candidate. We use an accuracy predictor to do this that obviates the need for 
network training. Two desirable properties of such a predictor are prediction reliability and 
sample efficiency. For prediction reliability, the predictor should minimize the distance between the 
predicted accuracy and the actual accuracy. In addition, it should maintain the relative ranking of 
architectures based on their predicted and actual accuracies. Furthermore, it should need a
small amount of training data (i.e., trained architectures).  We address these issues next.

\subsubsection{Regression Model}
\begin{figure*}[!ht]
    \centering 
    \includegraphics[scale=0.19]{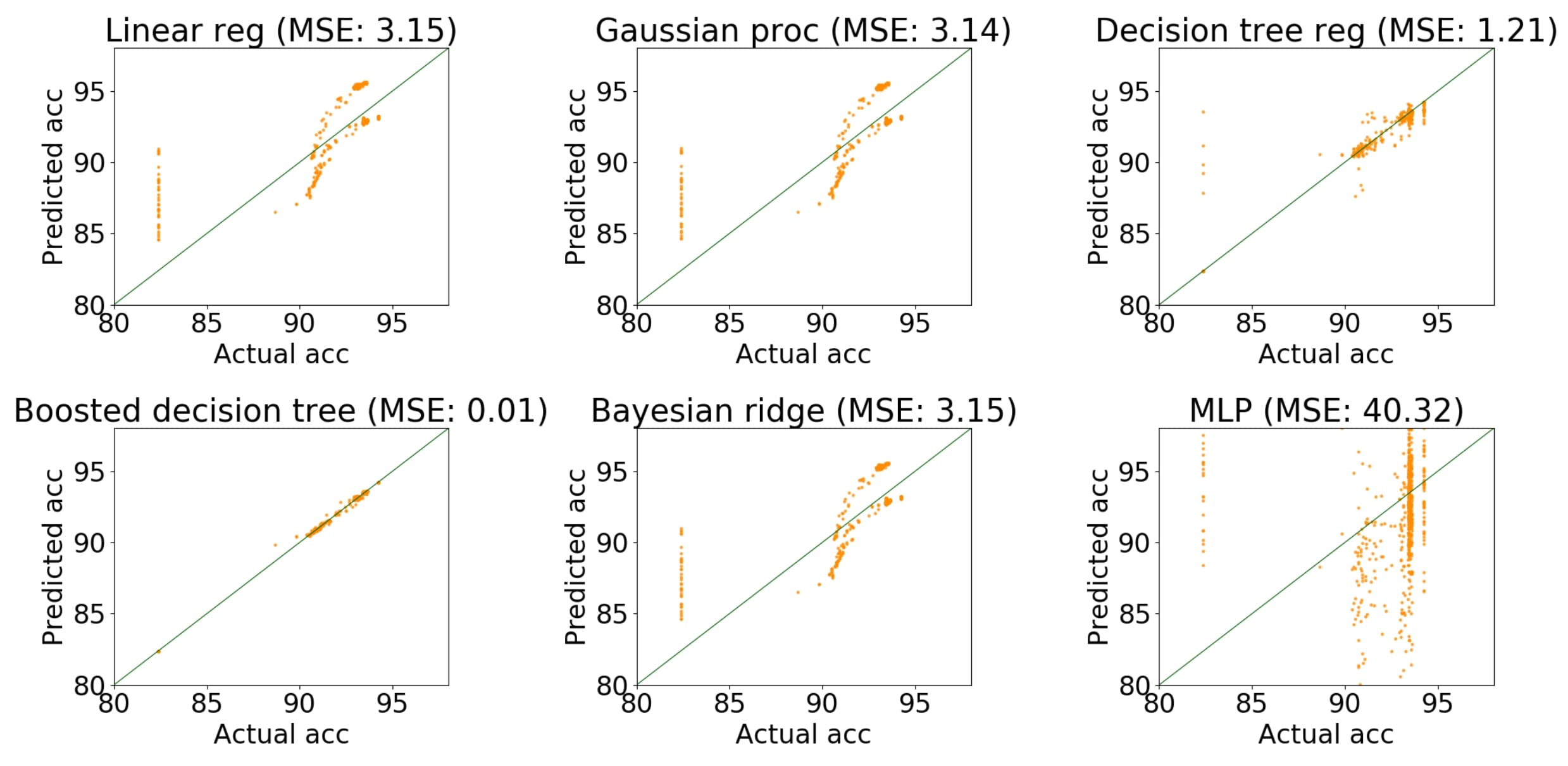}
    \caption{Performance comparison of six different regression methods for accuracy prediction.}
\label{fig:reg}
\end{figure*}

In order to obtain a suitable regression model for use in the accuracy predictor, we compared 
six methods. These are Gaussian process, multilayer perceptron (MLP), linear regression, 
decision tree regression, boosted decision tree regression, and Bayesian ridge regression. 
We compared their performance in the context of accuracy prediction. Fig.~\ref{fig:reg} shows a 
comparison of these methods based on $300$ sampled architectures (sample selection is discussed
ahead). The mean-squared error (MSE) is based on leave-one-out MSE. Since 
boosted decision tree regression had the smallest MSE, we chose it as the regression model for 
the accuracy predictor. 

Boosting is an ensemble approach that fits several copies of the regressor model on the original 
dataset. However, after each iteration, the weights of data instances are modified based on the 
prediction error of the current iteration. This process helps the regressor to focus more on 
difficult data instances in the next iteration. This is one reason why the boosted decision
tree regressor is very suitable for training the accuracy predictor. We set the maximum depth of the 
tree to five and the maximum number of estimators in the boosting stage to $500$. 

\subsubsection{Sample Generation}
The objective of this step is to generate a sample set that is representative of the architectures in 
the search space and has as few samples as possible. The process of iterative sample selection is 
summarized in Algorithm \ref{alg:sample}. 

Since the search space can be quite large, we first create a pool of architectures. 
We use quasi Monte-Carlo (QMC) sampling \cite{asmussen2007stochastic} methods, Sobol sequence 
\cite{sobol1967distribution} in particular, to generate the initial pool of architectures. 
Sobol sequences are designed to generate samples to be as uniformly distributed as possible over 
the multi-dimensional search space. In addition, compared to other sampling methods, such as random 
sampling, stratified sampling, and Latin hypercube sampling \cite{stein1987large}, sampling based on 
Sobol sequences generates more evenly distributed samples \cite{burhenne2011sampling}.  
Our initial pool of architectures contains $2048$ sample architectures.

After generating the pool of candidate architectures, we select a set of sample architectures to be 
trained and used for building the accuracy predictor.  This is done over multiple iterations (in our 
implementation, we used three iterations).  In the first step, a pre-defined number (set to $100$
in our experiments) of architectures are randomly selected from the pool of architectures. The 
accuracy predictor is 
trained based on \textit{architecture} accuracy obtained on a \textit{validation set}. Then, 
the fitness of the remaining samples in the architecture pool is evaluated based on the accuracy 
predictor. In the next iteration, we choose the samples with the highest accuracy prediction values. 
This process is repeated for a pre-defined number of iterations. In each iteration, the accuracy 
predictor gets updated with the new and updated set of 
\textit{architecture}-\textit{validation accuracy} observations. 

\begin{algorithm}[h]
    \caption{Iterative Sample Selection and Accuracy Predictor Training}
    \label{alg:sample}
    \begin{algorithmic}[l]
        \REQUIRE \textit{count}: Size of architecture pool; \textit{iterCount}: \#candidates in each 
iteration; $I_{max}$: \#iterations
        \STATE Lower/upper bounds and the step size of each hyperparameter
        \STATE \textit{archPool} = Sample $count$ Sobol sequence samples from the search space
        \STATE \textit{Samples} = Randomly select \textit{iterCount} samples from \textit{archPool}
        \STATE \textit{samplesAcc} = Validation-Acc(\textit{Samples})
        \STATE \textit{Predictor} = Boosted decision tree predictor trained with (\textit{Samples}, \textit{samplesAcc})
        \WHILE{maximum iterations $I_{max}$ not reached}
        \FOR{\textit{Arch} in \textit{Pool} $\setminus$ \textit{Samples}}
        \STATE {\textbf{(a)}} $Acc_{i}$ = \textit{Predictor} (\textit{Arch})
        \ENDFOR
        \STATE {\textbf{($1$)}} \textit{Arch} = Reverse-Sort(\textit{Arch}) \COMMENT{based on $Acc_{i}$}
        \STATE {\textbf{($2$)}} \textit{Candidates} = \textit{Arch} [:\textit{iterCount}]
        \STATE {\textbf{($3$)}} \textit{Samples} = \textit{Samples} $\cup$ \textit{Candidates}
        \STATE {\textbf{($4$)}} \textit{candidatesAcc} = Validation-Acc(\textit{Candidates})
        \STATE {\textbf{($5$)}} \textit{Predictor} = Boosted decision tree predictor trained with all observations
        \ENDWHILE
        \ENSURE The accuracy predictor model
    \end{algorithmic}
\end{algorithm}

\subsection{Efficient Evolutionary Search}
\label{sect:EES}
The EES process is formulated as an optimization problem to find the architecture that scores
highest on the fitness criterion: predicted accuracy. We use the accuracy predictor described in 
Section \ref{sect:acc-pred} to accelerate the search process. 

Algorithm \ref{alg:search} summarizes the search process. We use a genetic search algorithm to find 
the best architecture in the search space. The first step is to define the search space by 
identifying the lower and upper bound values for each hyperparameter. We also have to identify the 
step size for the values within this range.  Then we randomly generate a pre-defined number of 
architectures within this space. We evaluate the fitness of these architectures using the accuracy 
predictor.  Using the mutation operator, with a pre-defined probability, we breed the next generation 
of NN architecture candidates. We sort these candidates based on their predicted accuracy values and 
based on the number of samples in each search iteration, we pick the best architectures for the next 
iteration.  After a certain number of search process iterations, we find the best architecture. 
This architecture is trained and then evaluated on the test set. Thus, this process returns the 
best-found architecture and its performance on the test set. Here, performance refers to the accuracy 
of the best architecture after the global search step.  In our implementation, we set the population 
size in each iteration to $100$.  The total number of search iterations is set to $200$.

\begin{algorithm}[h]
    \caption{Global Search Module}
    \label{alg:search}
    \begin{algorithmic}[l]
        \REQUIRE \textit{getAcc}: Accuracy prediction; \textit{iterCount}: \#samples in each 
iteration; $\rho$: mutation probability; $I_{max}$: \#search iterations
        \STATE Lower/upper bounds and the step size of each hyperparameter
        \STATE Parents ['Arch'] = Randomly generate $iterCount$ architectures
        \STATE Parents ['Reward'] = \textit{getAcc} (Parents['Arch'])
        
        \WHILE{maximum iterations $I_{max}$ not reached}
        \STATE \textbf{($1$)} Children = $\Phi$
        \FOR{parent in Parents}
        \STATE {\textbf{(a)}} child = parent
        \STATE {\textbf{(b)}} with $\rho$: child ['Arch'] = \textit{mutate} parent['Arch']
        \STATE {\textbf{(c)}} child['reward'] = \textit{getAcc}(child['Arch'])
        \STATE {\textbf{(d)}} Children += child
        \ENDFOR
        \STATE {\textbf{($2$)}} Children = Children $\cup$ Parents
        \STATE {\textbf{($3$)}} Reverse-sort(Children) \COMMENT{based on 'reward'}
        \STATE {\textbf{($4$)}} \textit{bestArch} = Children[$0$]['Arch']
        \STATE {\textbf{($5$)}} \textit{bestReward} = Children[$0$]['reward']
        \STATE {\textbf{($6$)}} Parents = Children[:$iterCount$]
        \ENDWHILE
        \STATE Accuracy = Test accuracy of \textit{bestArch}
        \ENSURE Best architecture (\textit{bestArch}) and its test accuracy
    \end{algorithmic}
\end{algorithm}

\subsection{Grow-and-prune Synthesis}
\label{sect:grow-and-prune}
We use a secondary step of NN grow-and-prune synthesis for two main reasons: (i) to find a model with 
an even better performance and (ii) to ensure model compactness and computational efficiency.  
However, the initial NN architecture provided to a grow-and-prune NN synthesis tool has a notable 
impact on the result.  Hence, identifying a good starting point is important.  In our case, we treat 
the architecture found by the global search module to be the initial point. Subsequent grow-and-prune 
synthesis is encapsulated in a local search module.  We discuss application of the above
approach to FFNNs and CNNs next.

\subsubsection{FFNNs}
We use the SCANN \cite{hassantabar2019scann} synthesis tool for local search of the feed-forward part 
of the architecture. SCANN can achieve very high compression rates, yet generate very accurate 
networks, when a DR method is employed.

\begin{algorithm}[h]
    \caption{Local search with SCANN}
    \label{alg:scann}
    \begin{algorithmic}[l]
        \REQUIRE Best architecture from global search module $A_{init}$; Reduced-dimension dataset 
$(X,y)$, and number of search iterations $I_{max}$
        
        \WHILE{$I_{max}$ not reached}
        \STATE {\textbf{(a)} Apply one architecture-changing operation\;}
        \STATE {\textbf{(b)} Train weights of the network on dataset ($X,y$) and test its performance 
on the validation set\;}
        \ENDWHILE
        
        \ENSURE Final network architecture $A_{final}$ with best performance on the validation set
    \end{algorithmic}
\end{algorithm}

Algorithm \ref{alg:scann} summarizes the grow-and-prune process for the FFNNs.  We use the best 
architecture found from the search process discussed in Section \ref{sect:EES} to initialize this 
process.  We also feed SCANN the corresponding reduced-dimension dataset.  Subsequently, for a 
defined number of iterations, we use one of the three architecture-changing SCANN operations and 
evaluate the performance of the resulting model on the validation set. These operations are 
connection growth, neuron growth, and connection pruning.  The output of this module is the best 
architecture based on evaluation on the validation set.

\subsubsection{CNNs}
For the convolutional layers of the architecture, we use NeST \cite{dai2017nest} to 
perform grow-and-prune synthesis.  In the growth phase, this methodology uses an intelligent feature 
map growth method. In order to add a new feature map to the convolutional layers, the best set of 
kernels is selected from a set of randomly generated kernels. This selection is based on 
which ones reduce the value of the loss function the most.  We also employ \emph{partial-area 
convolutions}. Different parts of the images are of interest to different kernels. Therefore, using 
the method in NeST, we identify the corresponding areas of interest in the image and prune away the 
connections to other image areas. This reduces the number of parameters in the convolutional layers 
(by a certain pruning ratio), as well as the number of floating-point operations (FLOPs) needed for 
computation.

\section{Experimental Results}
\label{sect:expresults}
In this section, we evaluate the performance of STEERAGE on several datasets. 
Table \ref{tab:characteristics} shows their characteristics. 

The evaluation results are divided into two parts.  Section \ref{sect:first} presents results 
obtained by STEERAGE on FFNNs.  Section \ref{sect:second} presents results on the CNN architectures 
for the MNIST and CIFAR-$10$ datasets.  Note that MNIST has both types of architectures: an
FFNN LeNet-300-100 and CNN LeNet-5.  We use various ResNet \cite{he2016deep, he2016identity} 
architectures as the baseline for the CIFAR-$10$ dataset.  We demonstrate that the NNs generated by 
STEERAGE are compact, computationally efficient, and accurate. Therefore, STEERAGE-generated NNs can 
be used in energy-constrained edge devices and IoT sensors.

\begin{table*}[htbp]
\caption{Characteristics of the datasets}
\label{tab:characteristics}
\centering
\begin{tabular}{lccccc}
\toprule
Dataset                                     & Training Set & Validation Set & Test Set & Features & Classes \\ \midrule
Sensorless Drive Diagnosis                  & $40509$      & $9000$         & $9000$   & $48$     & $11$   \\
Human Activity Recognition (HAR)            & $5881$       & $1471$         & $2947$   & $561$    & $6$     \\
Musk v$2$                                     & $4100$       & $1000$         & $1974$   & $166$    & $2$     \\
Pen-Based Recognition of Handwritten Digits & $5995$       & $1499$         & $3498$   & $16$     & $10$    \\
Landsat Satellite Image                     & $3104$       & $1331$         & $2000$   & $36$     & $6$     \\
Letter Recognition                          & $10500$      & $4500$         & $5000$   & $16$     & $26$    \\
Epileptic Seizure Recognition               & $6560$       & $1620$         & $3320$   & $178$    & $2$     \\
Smartphone Human Activity Recognition       & $6121$       & $153$          & $3277$   & $561$    & $12$    \\
DNA                                         & $1400$       & $600$          & $1186$   & $180$    & $3$     \\
MNIST                                       & $50000$      & $10000$        & $10000$  & $784$    & $10$    \\
CIFAR-$10$                                       & $55000$      & $5000$        & $10000$  & $3072$    & $10$    \\
\bottomrule
\end{tabular}
\end{table*}

\subsection{FFNNs}
\label{sect:first}

In this section, we present STEERAGE synthesis results on FFNNs. We started with a large set of 
potential hyperparameters and finalized it into a small subset. 
Table \ref{tab:hyperffnn} shows the hyperparameters we considered for FFNNs. 
For each hyperparameter, we show the lower bound, upper bound, and step size. 
For the number of layers, we considered one to six hidden layers in the architecture search space. 
with a step size of one.  We let the synthesis algorithm find the optimal number of neurons 
(in a 50-600 range, with a step size of 25) in each of the hidden layers.  STEERAGE steps
through the 11 DR method discussed earlier one by one, with the DR ratio spanning 1-20 with a
step size of 0.1.  Finally, we evaluated four different values for quantization: 4-, 8-, 16- or 
$32$-bit (i.e., full precision) inference. These are placed in bins 1-4 in the table.
Furthermore, we may refine these search parameters for specific datasets. 
We also considered use of different nonlinear activation functions in different network layers. 
However, since this did not help improve performance, we did not consider it further.

The final search space for each dataset is a subset of the search space defined by this table. 
For example, for the letter recognition dataset that only has $16$ features, we considered feature 
compression ratios up to $6\times$.  In the following, first, 
we present synthesis results on nine small- to medium-size datasets whose NN models would be 
appropriate for edge devices or IoT sensors.  These datasets were obtained from the UCI machine 
learning repository \cite{Dua:2017} and Statlog collection \cite{Michie:1995:MLN:212782}.
Then, we present results on the FFNN implementation of the MNIST dataset.  

\begin{table}[]
\caption{General search space for FFNNs}
\label{tab:hyperffnn}
\begin{tabular}{llll}
\toprule
Hyperparameter                  & Lower bound & Upper bound & Step size \\ \midrule
\#Layers                        & $1$           & $6$           & $1$    \\ 
\#Neurons per layer             & $50$          & $600$         & $25$   \\ 
DR method & $1$           & $11$          & $1$   \\ 
DR ratio  & $1$           & $20$          & $0.1$  \\ 
Quantization                    & bin $1$           & bin $4$           & $1$    \\ 
\bottomrule
\end{tabular}
\end{table}

\subsubsection{Small to medium size datasets}
The top nine rows of Table \ref{tab:characteristics} 
show the characteristics of the small- 
to medium-size datasets we experimented with.  For these experiments, we use the Adam optimizer with 
a learning rate of $0.01$ and weight decay of $1$e-$3$.  Table \ref{tab:hyper-ffnn-acc} shows the 
test accuracy results.   There are two rows associated with STEERAGE.  The first one shows the result 
obtained by just using the global search module, and the second one when both the global and local 
search modules are used.  The number of neurons in the hidden layers of the baseline MLP architecture 
is a multiple of the number of its input features.  Relative to the MLP baseline, STEERAGE-generated 
NNs improve classification accuracy by $0.51\%$ to $10.19\%$.  Furthermore, compared to SCANN 
and DR+SCANN \cite{hassantabar2019scann}, STEERAGE generates NNs with higher accuracy across all 
datasets.  Table \ref{tab:hyper-ffnn-param} shows the number of parameters in the respective 
NN architectures in Table \ref{tab:hyper-ffnn-acc}.  As can be seen, STEERAGE, in general,
generates much more compact architectures than the baseline and is competitive with SCANN. 
In addition, using the global+local search helps both in terms of model performance 
and connection compression ratio relative to just global search.

\begin{table*}[t]
\caption{Highest test accuracy comparison on small- to medium-size datasets (the highest number is 
highlighted)}
\label{tab:hyper-ffnn-acc}
\centering
\begin{tabular}{lccccccccc} 
\toprule
Dataset                      & SenDrive  & HAR       & Musk      & Pendigits & SatIm     & Letter    & Seizure   & SHAR      & DNA       \\ \midrule
Baseline                     & $93.53\%$ & $95.01\%$ & $98.68\%$ & $97.22\%$ & $91.30\%$ & $95.24\%$ & $87.53\%$ & $90.66\%$ & $94.86\%$ \\
SCANN \cite{hassantabar2019scann}                      & $97.10\%$ & $95.52\%$ & $99.09\%$ &
$97.22\%$ & $90.10\%$ & $92.60\%$  & $96.96\%$ & $93.78\%$ & $95.86\%$ \\
DR + SCANN \cite{hassantabar2019scann}                  & $99.34\%$ & $95.28\%$   & $98.08\%$ & $97.93\%$ & $89.40\%$ & $92.70\%$ & $97.62\%$ & $94.84\%$ & $93.76\%$ \\ \hdashline
STEERAGE (GS)         & $99.07\%$   & $95.90\%$  & $98.90\%$ & $97.30\%$ & $90.35\%$ &
$\textbf{97.20\%}$  & $95.50\%$  & $94.86\%$ & $95.56\%$ \\
STEERAGE (GS+LS) &     $\textbf{99.36\%}$  & $\textbf{96.43\%}$ & $\textbf{99.19\%}$ &
$\textbf{98.05\%}$ & $\textbf{92.00\%}$ & $95.10\%$  & $\textbf{97.72\%}$ & $\textbf{95.50\%}$  & $\textbf{95.95\%}$ \\ \bottomrule
\end{tabular}
\end{table*}

\begin{table*}[t]
\caption{Neural network parameter comparison}
\label{tab:hyper-ffnn-param}
\centering
\begin{tabular}{lccccccccc} 
\toprule
Dataset                      & SenDrive  & HAR       & Musk      & Pendigits & SatIm     & Letter    & Seizure   & SHAR      & DNA       \\ \midrule
Baseline                     & $56.9$k   & $212.0$k   & $55.8$k    & $4.9$k    & $3.8$k   & $4.4$k   & $380.9$k   & $214.0$k    & $24.6$k   \\
SCANN                        & $10.0$k  & $5.0$k  & $22.0$k  & $3.2$k  & $3.2$k  & $3.8$k & $3.0$k    & $10.0$k  & $20.0$k  \\
DR + SCANN                   & $2.2$k  & $1.0$k  & $600$    & $400$  & $1.0$k    & $3.7$k & $1.8$k  & $1.0$k   & $300$                       \\ \hdashline
STEERAGE (GS)       & $15.8$k                              & $48.1$k                             & $38.7$k                            & $4.9$k                           & $3.2$k                           & $4.0$k                            & $86.8$k                              & $20.9$k                             & $20.3$k                            \\

STEERAGE (GS+LS) & $5.0$k                            & $6.5$k                            &
$10.0$k                            & $2.0$k                           & $3.5$k
& $10.0$k                            & $10.0$k                             & $5.0$k                          & $5.0$k  \\                         
\bottomrule
\end{tabular}
\end{table*}

\subsubsection{The MNIST dataset}
MNIST is a well-studied dataset of handwritten digits. It contains 60000 training images and 10000
test images. We set aside 10000 images from the training set as the validation set. We adopt
LeNet-$300$-$100$ \cite{lecun1998gradient} as the baseline FFNN model for this dataset.
We use the stochastic gradient decent optimizer in our experiments, with a learning rate of
$0.01$, momentum of $0.9$, and weight decay of $1$e-$4$.  Table \ref{tab:mnist-ffnn} shows the
results for feed-forward architecture synthesis for the MNIST dataset.
We also compare the results of our synthesis methodology with other related work on
feed-forward architectures on the MNIST dataset.
Relative to related work, the combination of global and local search achieves
the highest accuracy of $99.26\%$, with a connection compression ratio of $10.2\times$. 
In this case, DR+SCANN generates the most compact NNs, however, at a lower accuracy than
STEERAGE.  This points to a classic accuracy-compactness tradeoff.  Again, using the combination 
of global and local search helps both in terms of model performance and connection compression ratio.

\begin{table*}[!htbp]
\caption{Accuracy results for feed-forward MNIST architectures}
\label{tab:mnist-ffnn}
\centering
\begin{tabular}{llll}
\toprule
Method & Weights  & Connection Compression ratio & Test Accuracy \\ \midrule
LeNet-$300$-$100$ & $266$k & $1.0\times$ & $98.70\%$\\
NeST \cite{dai2017nest} & $7.8$k & $34.1\times$ &  $98.71\%$ \\
DR + SCANN  \cite{hassantabar2019scann} & $23$k & $11.6\times$ & $99.24\%$\\
DR + SCANN \cite{hassantabar2019scann} & $2.5$k & $106.4\times$ & $98.73\%$\\ \hdashline
STEERAGE (GS) & $135$k & $2.0\times$ & $99.16\%$\\
STEERAGE (GS+LS) & $26$k & $10.2\times$ & $99.26\%$\\
\bottomrule
\end{tabular}
\end{table*}

\subsection{CNNs}
\label{sect:second}
In this section, we present results of STEERAGE on CNN architectures obtained for 
the medium-size MNIST and CIFAR-$10$ datasets. 

\subsubsection{The MNIST dataset}
For the MNIST dataset, we use LeNet-$5$ \cite{lecun1998gradient} as the baseline architecture. We 
use the PyTorch \cite{paszke2017automatic} implementation of this network.  This architecture has two 
convolutional layers with $6$ and $16$ filters, respectively.  The convolutional layers are followed 
by a feed-forward classifier with three hidden layers consisting of $400$, $120$, and $84$ neurons, 
respectively.  Table \ref{tab:lenet5-search} shows the search space. The search space includes 
image DR method, image size, and quantization.  It also includes the number of filters in each 
convolutional layer (between $5$ to $10$ for the first layer, and $15$ to $20$ for the second one), 
use of max or average pooling after the convolutional layers, and the number of neurons in the three 
fully-connected layers. The search space for the fully-connected layers is selected 
based on image size. As the number of input features decreases, we decrease the search interval 
for the number of neurons in these layers. 

\begin{table}[ht]
\caption{Search space for LeNet-$5$}
\label{tab:lenet5-search}
\centering
\begin{tabular}{llll}
\toprule
Input size & \begin{tabular}[c]{@{}l@{}}$14\times 14$\\ $(196)$\end{tabular} & \begin{tabular}[c]{@{}l@{}}$21 \times 21$\\ $(441)$\end{tabular} & \begin{tabular}[c]{@{}l@{}}$28 \times 28$\\ $(784)$\end{tabular} \\ \midrule
Image DR  & \multicolumn{3}{c}{$1$ - $6$}  \\ \midrule
Kernel Size & \multicolumn{3}{c}{$3\times3$ vs. $5\times5$} \\ \midrule
Conv$1$    & $5$ - $10$                                                    & $5$ - $10$                                                     & $5$ - $10$                                                     \\
Pooling$1$  & \multicolumn{3}{c}{Max vs. Avg}  \\
Conv$2$    & $15$ - $20$                                                   & $15$ - $20$                                                    & $15$ - $20$                                                    \\
Pooling$2$  & \multicolumn{3}{c}{Max vs. Avg}  \\
FC$1$      & $135$ - $180$                                                 & $375$ - $500$                                                    & $540$ - $720$                                                    \\
FC$2$      & $60$ - $120$                                                  & $100$ - $220$                                                  & $100$ - $250$                                                    \\
FC$3$      & $40$ - $85$                                                    & $50$ - $120$                                                     & $50$ - $150$\\ \midrule
Quantization bits & \multicolumn{3}{c}{$4$ - $8$ - $16$ - $32$}  \\\bottomrule                                
\end{tabular}
\end{table}

We use the stochastic gradient descent optimizer in our experiments, with a learning rate
of $0.001$, momentum of $0.9$, and weight decay of $1$e-$4$ in the global search module.
In the local search module, we use various SCANN schemes to obtain the feed-forward part of the 
network and NeST for grow-and-prune synthesis of the convolutional layers. 

Table \ref{tab:mnist-cnn} presents the results.  The most accurate architecture based on 
the LeNet-$5$ model has an error rate of $0.66\%$. It is based on using both global and local search 
modules and corresponds to images of size $21 \times 21$.  This model only has $7.2$k parameters. 
Moreover, another model based on global+local search only has $5.2$k parameters, with an error rate 
of $0.72\%$.  These two results are highlighted in bold.  Interestingly, the first of the above two 
models outperforms the GS+LS model based on an image size of $28 \times 28$.  This may be due
to the fact that dimensionality reduction has two opposite effects on accuracy: mitigation of
curse of dimensionality improves accuracy; however, it also results in some loss of information
that reduces accuracy.

\begin{table}[]
\caption{Comparison of results for LeNet-$5$ (the size of the image is shown in parentheses
for STEERAGE)} 
\label{tab:mnist-cnn}
\centering
\begin{tabular}{lll}
\toprule
Method                                & Error rate (\%)
 & Weights  \\ \midrule
Baseline                               & $0.80$      & $62.0$k    \\
Network pruning  \cite{han2015learning}                     & $0.77$     & $34.5$k  \\
NeST \cite{dai2017nest}                                   & $0.77\%$   & $5.8$k   \\
SCANN Scheme A\cite{hassantabar2019scann}                                 & $0.68\%$   & $186.4$k \\
SCANN Scheme C\cite{hassantabar2019scann}                                & $0.72\%$   & $9.3$k   \\ \hdashline
STEERAGE (GS) ($14 \times 14$)         & $4.96\%$   & $38.9$k  \\ 
STEERAGE (GS+LS) ($14 \times 14$) & $1.2\%$    & $5.0$k   \\
STEERAGE (GS) ($21 \times 21$)         & $1.9\%$    & $207.2$k \\
\textbf{STEERAGE (GS+LS) ($21 \times 21$)} & $\textbf{0.72\%}$   & $\textbf{5.2}$\textbf{k}   \\
\textbf{STEERAGE (GS+LS) ($21 \times 21$)} & $\textbf{0.66\%}$   & $\textbf{7.2}$\textbf{k} \\
STEERAGE (GS) ($28 \times 28$)         & $0.86\%$   & $131.1$k \\
STEERAGE (GS+LS) ($28 \times 28$) & $0.68\%$   & $9.9$k \\ \bottomrule
\end{tabular}
\end{table}

\begin{figure*}[!ht]
    \centering
    \subfloat[]{\includegraphics[scale=0.25]{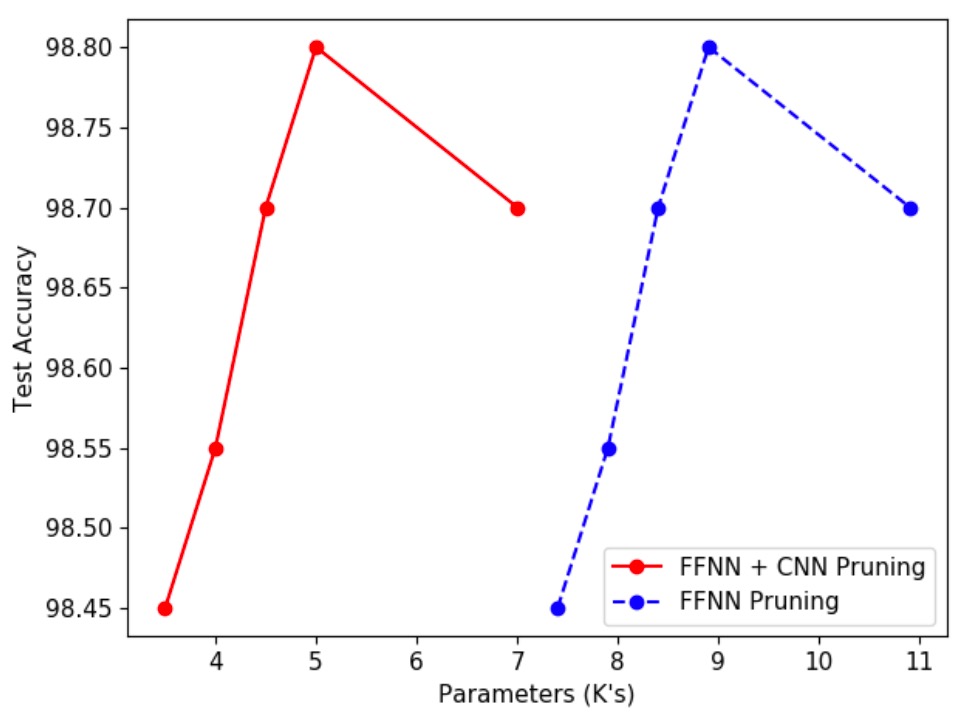}}
    \label{fig:1a}
    \hfil
    \subfloat[]{\includegraphics[scale=0.25]{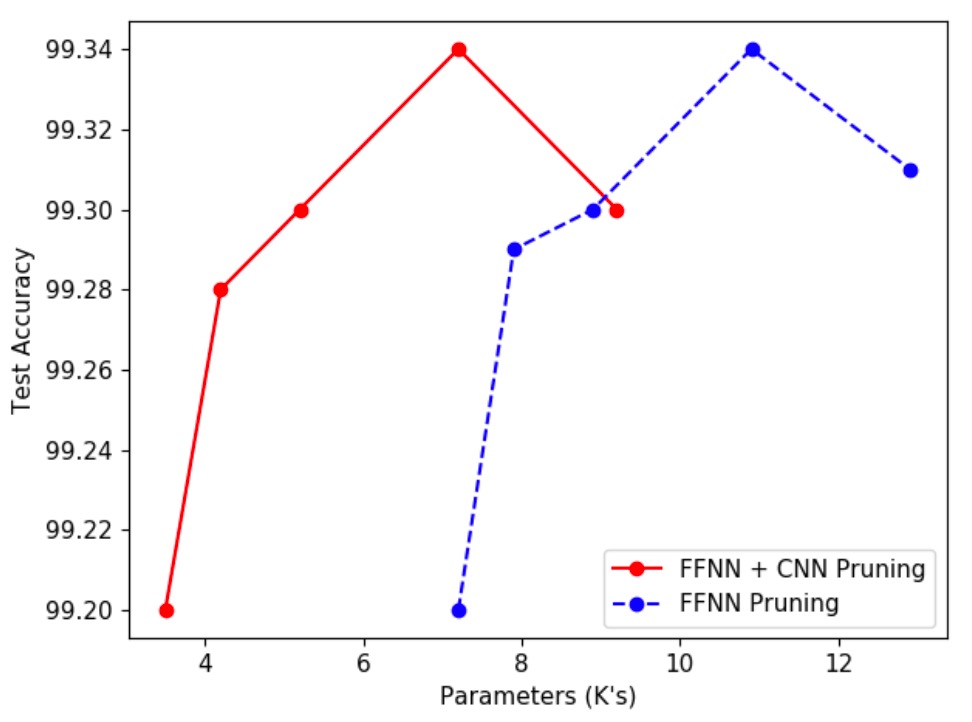}}
    \label{fig:1b}
    \subfloat[]{\includegraphics[scale=0.25]{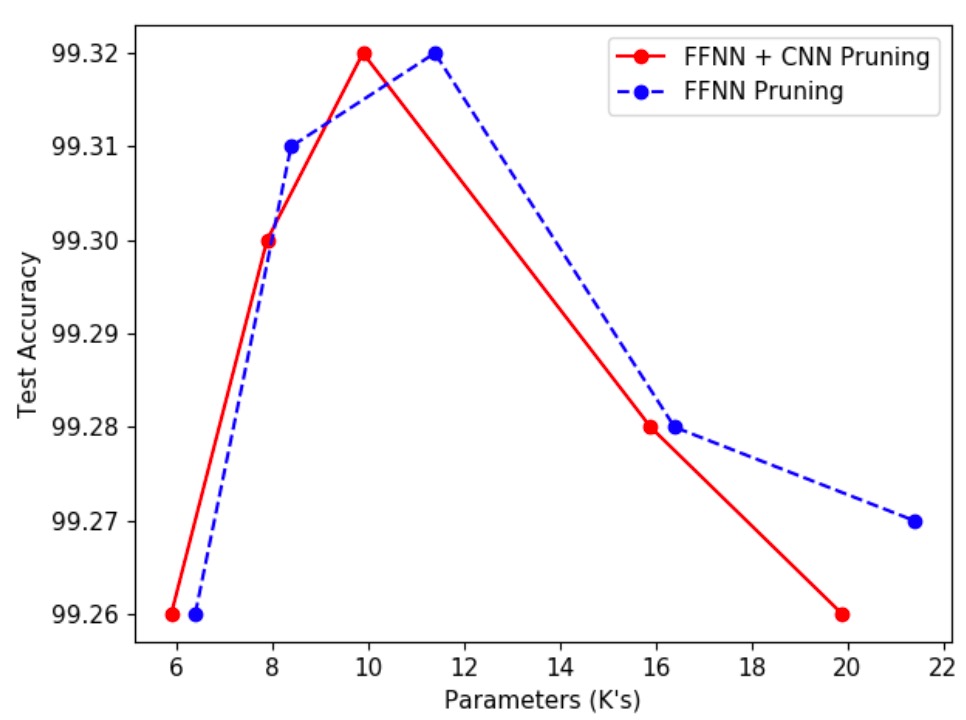}}
    \label{fig:1c}
    \caption{Accuracy vs.~\#parameters tradeoff for three variants of LeNet-$5$ for images sizes: 
(a) $14\times14$, (b) $21\times21$, and (c) $28\times28$.}
    \label{fig:lenet5-tradeoff}
\end{figure*}

Fig.~\ref{fig:lenet5-tradeoff} shows the tradeoff between accuracy and the number of parameters in 
the network for various image sizes.  For each image size, we show two graphs.  The first is the 
result of using grown-and-prune synthesis on only the feed-forward part of the architecture,
whereas the other graph is based on using it on both the feed-forward and convolutional layers.
As can be seen, using it on both layer types yields NNs that are both more accurate and more 
compact.  Furthermore, by using global and local search modules, STEERAGE provides the NN designers 
a choice among NNs that fall at different points on the graph and is most suitable to their use
scenario.  Note that beyond a certain point, adding more parameters to the network leads to the 
problem of overfitting and thus damages performance. 

\subsubsection{The CIFAR-$10$ dataset}
In this section, we present results of our CNN synthesis methodology for the CIFAR-$10$ dataset.
For this dataset, we used ResNet \cite{he2016deep, he2016identity} architectures of various
depths as the baselines.  ResNet uses two residual block architectures, basic and bottleneck, to 
facilitate training of deep networks.  The search space for ResNet architectures is shown in Table 
\ref{tab:cifar-hyper}.  We used two different image sizes: the original $32\times32$ and 
down-sampled $24\times24$.  We choose from among the six DR methods discussed earlier for images. 
ResNet architectures have four stages of residual blocks, either basic or bottleneck 
\cite{he2016deep}.  For ResNet architectures that are less than (greater than equal to) $50$ layers 
deep, we use the basic (bottleneck) block.  We search for architectures that are between $18$ to 
$50$ layers deep (in steps of $2$ layers), by searching for the number of basic blocks (in the
range 2-6) for each of the four stages in the residual network.  Moreover, the number of filters in 
the convolutional layers of each stage is part of the search space.  Other hyperparameters 
denote whether max or average pooling is used after the convolutional layers, whether the
number of fully-connected layers is $1$ or $2$, and the quantization level (4, 8, 16 or 32 bits
of precision).  We also experiment with fixed-depth ResNet architectures with $18$ layers. In these 
experiments, we do not include network depth in the search space.

\begin{table}[]
\caption{Adaptive search space for ResNet architectures (for \#filters in the convolutional layers, 
the step sizes are shown in the parentheses)} 
\label{tab:cifar-hyper}
\centering
\begin{tabular}{lc} 
\toprule
Input size     & $(3\times24\times24)$ - $(3\times32\times32)$ \\ \midrule
Image DR       & $1$ - $6$                                           \\ \midrule
\#Basic blocks & $2$ - $6$                                             \\
Network depth  & $18$ - $50$ ($2$)                                   \\ \midrule
Conv$1$        & $56$ - $98$ ($8$)                                   \\
Conv$2$\_x     & $56$ - $98$ ($8$)                                   \\
Conv$3$\_x     & $112$ - $196$ ($16$)                                  \\
Conv$4$\_x     & $224$ - $392$ ($32$)                                  \\
Conv$5$\_x     & $448$ - $784$ ($64$)                                  \\
Pooling        & Max - Avg                                         \\ 
\#FC layers    & $1$ - $2$                                         \\ \midrule
Quantization bits   & $4$ - $8$ - $16$ - $32$                      \\ \bottomrule
\end{tabular}
\end{table}

The results are summarized in Table \ref{tab:cifar-result}. We compare the results obtained
using our synthesis methodology with those for the original ResNet architectures.  By setting different pruning ratios (in partial-area convolution operation) in the local search module, STEERAGE can navigate the accuracy-model complexity 
tradeoff. This is evident from the results shown in the table.  Our ResNet architecture variants are 
more accurate compared to the original ones. For example, STEERAGE-synthesized ResNet-$18$ can reach 
an error rate of just $4.46\%$, which is $2.52\%$ lower than the error rate of the original
ResNet-$18$, and $1.74\%$ lower than that of even the much larger ResNet-$101$.  Furthermore, another 
synthesized ResNet-$18$ with an error rate of $4.78\%$ has $2.1\times$ fewer parameters 
relative to the baseline ResNet-$18$. 

By including the depth of the network as one of the hyperparameters in the search space, we 
were able to synthesize a variant of the ResNet architecture with $40$ layers that can achieve 
an error rate of only $3.86\%$, with $0.41$M network parameters.  It dominates the original
ResNet-101 architecture in the error rate, number of parameters, as well as FLOPs. 
Another variant of this architecture with an error rate of $4.30\%$ reduces the number of
parameters and FLOPs further, and also dominates the original ResNet-101 architecture.
These STEERAGE-synthesized architectures are more accurate compared to even the ResNet-$1001$ 
architecture \cite{he2016identity}. ResNet-$1001$ has 1001 layers and is the most accurate ResNet 
architecture, with an error rate of $4.62\%$. 
This shows that the traditional method of increasing accuracy by increasing the number of layers is
not necessarily the best approach -- a better architecture can accomplish the same job with
much fewer layers (as well as fewer parameters and FLOPs).

\begin{table}[]
\caption{Results for the CIFAR-$10$ dataset.} 
\label{tab:cifar-result}
\centering
\begin{tabular}{lccc}
\toprule
Method & Error rate (\%) & \#Parameters & FLOPs \\ \midrule
ResNet-$18$ \cite{he2016deep}       & $6.98$      & $0.25$M      &    $585.1$M   \\
ResNet-$44$ \cite{he2016deep} & $6.38$ & $0.66$M   &  $2.47$G   \\
ResNet-$101$    \cite{he2016deep}   & $6.20$    & $44.5$M      &  $2.47$G     \\ 
ResNet-$1001$ \cite{he2016identity} & $4.62$ & $10.2$M & $2.47$G \\ \hdashline
ResNet-$18$ (GS)       & $4.88$   & $0.36$M     &    $861.1$M   \\
ResNet-$18$ (GS+LS)     & $4.46$    & $0.27$M      &    $645.8$M   \\
ResNet-$18$ (GS+LS)     & $4.78$    & $0.12$M      &   $269.1$M    \\
ResNet-$40$ (GS)       & $4.35$   & $0.55$M      &    $1.80$G   \\
ResNet-$40$ (GS+LS)     & $3.86$   & $0.41$M      &   $1.30$G   \\
ResNet-$40$ (GS+LS)     & $4.30$   & $0.21$M      &   $684.0$M    \\ \bottomrule
\end{tabular}
\end{table}

\section{Discussion}
\label{sect:discussion}
There are several advantages to our synthesis methodology. 
To begin with, we speed up the search process by using
the accuracy predictor models.  
For example, training a ResNet-$18$ architecture for $200$ epochs can take around $2$ GPU hours. 
However, by using the accuracy predictor, we can evaluate this architecture in $1.2$ seconds. This results in an acceleration rate of around $6000\times$ in evaluating the architectures in the search process.
Furthermore, by using the boosted decision tree regression, we trained a predictor that accurately predicts the performance of the architectures.
This synthesis approach is more efficient relative to RL-based NAS approaches. 
The methodology also easily adapts to various network architecture types: FFNNs 
as well as shallow and deep CNNs. 

Our architecture search framework is general. As a result, we can easily add other hyperparameters to 
the search space. For example, an interesting topic for future research is to investigate the effect 
of including the type of normalization on architecture performance.  Similar to the work done in 
\cite{luo2019switchable}, we could add switchable normalization to the search space and 
enable the framework to find the optimal combination of various normalization schemes for different 
parts of the network. 

We evaluated the impact of both global search and combined global+local search.
The combined approach performed better.  This is because global search efficiently obtains a good 
initialization point for subsequent grow-and-prune synthesis that yields a compact and
computationally-efficient architecture while also enhancing accuracy.

\section{Conclusion}
\label{sect:conclusion}
In this article, we proposed a new two-step NN synthesis methodology called STEERAGE. It uses an
efficient global search module based on an accuracy predictor to find the best set of
hyperparameter values for the NN design space. It then uses grow-and-prune synthesis methods to 
find a superior version of the architecture that is more compact, efficient, and accurate.
Experimental results demonstrated the ability of STEERAGE to generate accurate and
compact networks for both architecture types: FFNNs and CNNs. STEERAGE generated a ResNet-$40$
architecture that achieves the highest accuracy relative to all ResNet architectures on 
the CIFAR-$10$ dataset.
\bibliographystyle{IEEEtran}
\bibliography{STEERAGE}

\end{document}